\documentclass{article}

\usepackage[table]{xcolor}
\usepackage[final]{corl_2026}

\makeatletter
\renewcommand{\@notice}{}
\makeatother

\usepackage{amsmath}
\usepackage{amssymb}
\usepackage{booktabs}
\usepackage{graphicx}

\usepackage{pifont}
\usepackage{placeins}
\usepackage{enumitem}

\usepackage{array}
\usepackage{longtable}
\usepackage{multirow}
\usepackage{wrapfig}

\newcommand{\modelname}{AHA-WAM}
\newcommand{\cmark}{\ding{51}}
\newcommand{\xmark}{\ding{55}}

\definecolor{bestgreen}{RGB}{226, 246, 226}
\definecolor{secondgreen}{RGB}{242, 251, 242}

\title{AHA-WAM:Asynchronous Horizon-Adaptive World-Action Modeling with Observation-Guided Context Routing}

\author{
  \bfseries Jisong Cai$^{1,2 *}$ \quad Long Ling$^{1,3*}$ \quad Shiwei Chu$^{1}$ \quad Zhongshan Liu$^{3}$\\
  \bfseries Jiayue Kang$^{1}$ \quad Zhixuan Liang$^{4,2}$ \quad Wenjie Xu$^{3}$ \quad Yinan Mao$^{3}$\\
  \bfseries Weinan Zhang$^{1,2}$ \quad Xiaokang Yang$^{1}$ \quad Ru Ying$^{3}$ \quad Ran Zheng$^{3}$ \quad Yao Mu$^{1,2 \dagger}$\\[5pt]
  \normalfont\normalsize $^{1}$Shanghai Jiao Tong University \quad $^{2}$Shanghai AI Laboratory\\[-1pt]
  \normalfont\normalsize $^{3}$Baidu AI Cloud \quad $^{4}$The University of Hong Kong\\[12pt]
  \normalfont\normalsize \texttt{Project Page}: \href{https://serene-sivy.github.io/aha-wam/}{\texttt{https://serene-sivy.github.io/aha-wam/}}\\[-6pt]
}

\begin{document}
\maketitle
\begingroup
\renewcommand{\thefootnote}{\fnsymbol{footnote}}
\footnotetext[1]{Equal contribution.}
\footnotetext[2]{Corresponding author. Email: \texttt{muyao@sjtu.edu.cn}.}
\begingroup
\makeatletter
\renewcommand{\@makefnmark}{}
\footnotetext[3]{This work was done during Long Ling's and Jiayue Kang's internship at Shanghai Jiao Tong University.}
\makeatother
\endgroup
\endgroup
\vspace{-0.15in}

\begin{abstract}
World-action models have emerged as a promising paradigm for robot manipulation, jointly modeling visual scene dynamics and actions to inject physical priors into policy learning.
However, existing world-action models couple world prediction and action execution at the same temporal resolution, forcing the world branch to model near-term frame variations that are redundant and weakly informative.
We posit that strictly binding world prediction and action execution to the same temporal rhythm may underutilize the potential of the video branch for embodied control.
Therefore, we propose \modelname{}, an \textbf{A}synchronous \textbf{H}orizon-\textbf{A}daptive World-Action Model built on a dual Diffusion Transformer (DiT) architecture that reorganizes world-action modeling around this temporal asymmetry.
\modelname{} instantiates the video DiT as a low-frequency world planner that maintains rolling key-value memory over past observations and exposes reusable layerwise latent context encoding long-horizon scene evolution, while a high-frequency action DiT executes short action chunks in closed loop by querying this context through layerwise joint attention.
To support asynchronous execution, we introduce horizon-adaptive offset training and Observation-Guided Video-Context Routing (OVCR), which together let the action expert exploit long-horizon world context while remaining responsive to real-time execution state without rerunning the video DiT.
Experiments on RoboTwin and real-world manipulation tasks show that \modelname{} achieves state-of-the-art performance without any robot-data pretraining, attaining \textbf{92.80\%} average success on RoboTwin and \textbf{78.3\%} success across 4 real-world tasks, while reaching \textbf{24.17} Hz closed-loop control with a \textbf{4.59}$\times$ speedup over Fast-WAM.

\end{abstract}

\keywords{Robot Learning, Embodied Manipulation, World-Action Model}

\section{Introduction}

\begin{figure}[t]
    \centering
    \includegraphics[width=0.99\linewidth]{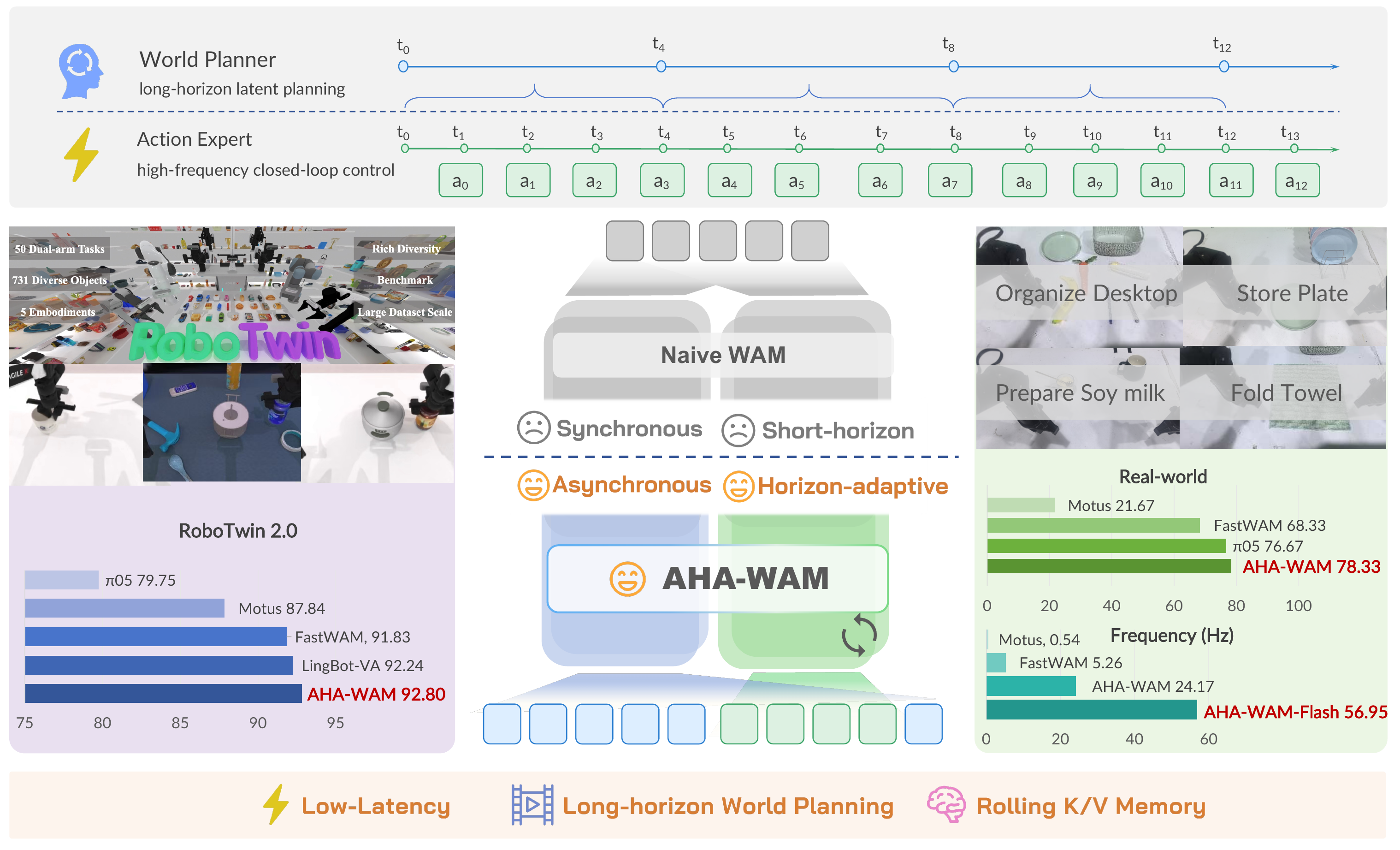}
    \vspace{-4pt}
    \caption{\textbf{Overview of \modelname{}.} \modelname{} connects past observations, future-oriented world planning, and fast closed-loop action execution: a slow world planner maintains reusable memory and planning context, while a fast action expert adapts that context to the latest observation before predicting short action chunks.}
    \label{fig:teaser}
    \vspace{-8pt}
\end{figure}

Robotic manipulation requires policies that understand not only the current scene, but also how the scene may evolve under the robot's actions.
Recent vision-language-action (VLA) models have advanced robot control by scaling imitation learning with large vision-language model backbones, yet action labels provide relatively sparse supervision for the underlying physical dynamics of manipulation.
World-action models (WAMs) address this by coupling action prediction with dense video-based world modeling, learning how actions co-evolve with visual scene dynamics to inject physical priors into control and offer a more generalizable, and transferable policy representation.

Despite this promise, current WAMs leave an important design space underexplored: how can the video branch more effectively empower robot control?
Existing approaches either explicitly roll out future frames for inverse-dynamics decoding or joint world-action modeling, while others reuse the video branch as a latent encoder alongside the action branch. Both designs assume that world prediction and action execution share the same short horizon, which forces the world branch to spend capacity on dense adjacent-frame variations that are often highly correlated and only weakly informative for control.
We argue that this coupling reflects a structural mismatch in temporal abstraction: the video world model can better serve embodied control by forming a temporally extended latent plan over future visual states, whereas the action model should remain tightly coupled to the real-time control loop, incorporating the latest observation to issue timely closed-loop corrections.

We propose \modelname{}, an \textbf{A}synchronous \textbf{H}orizon-\textbf{A}daptive World-Action Model that reorganizes world-action modeling around this temporal asymmetry. \modelname{} instantiates the video Diffusion Transformer (DiT) as a low-frequency world planner that maintains a rolling key-value (KV) memory over past observations and exposes reusable layerwise latent context encoding long-horizon scene evolution, amortizing expensive world-model computation across multiple action steps. In parallel, a high-frequency action DiT executes short action chunks in closed loop by querying this context through layerwise joint attention. This design preserves the benefits of joint world-action modeling while aligning each branch with the temporal scale at which it's most informative. In \modelname{}, \emph{horizon-adaptive} refers not to online horizon adjustment, but to a horizon-decoupled formulation where the world and action branches are assigned different temporal horizons according to their functional roles.

A key challenge introduced by asynchronous execution is that the planner context may become stale or misaligned with the current action chunk as the executor runs ahead.
To address this, we introduce Observation-Guided Video-Context Routing (OVCR), which constructs chunk-specific latent queries from the latest visual observations to route and update the planner's layerwise video context before it is consumed by the action DiT, giving each action prediction an observation-conditioned view of the slow planner state without feeding dense visual tokens into the high-frequency action branch.
Proprioceptive feedback enters the action DiT directly, while visual feedback is injected indirectly through the routed video context, keeping the reused planner context responsive to real-time execution state without recomputation.
We further introduce horizon-adaptive offset training to expose the model to diverse planner-executor phase relationships, and equip the video DiT with rolling KV memory to connect past observations with future plans.

We evaluate \modelname{} on RoboTwin 2.0 and real-world manipulation tasks.
Though without robot-data pretraining, \modelname{} achieves 92.80\% average success across 50 RoboTwin 2.0 tasks, reaching state-of-the-art-level performance among strong VLA and WAM baselines.
Across four real-world tasks covering deformable manipulation, long-horizon organization, fine-grained tool use, and spatial generalization, \modelname{} achieves 78.3\% average success, demonstrating robust deployment performance beyond simulation. Across four dimensions of real-world out-of-distribution evaluation, \modelname{} matches $\pi_{0.5}$ in exhibiting the most limited performance degradation, suggesting stronger robustness to distribution shifts.
Benefiting from the asynchronous inference schedule, ODE distillation and cuda optimizations, \modelname{} reaches up to 56.9 Hz closed-loop control frequency, corresponding to a 10.82$\times$ speedup over Fast-WAM.

In summary, our contributions are threefold:

\begin{itemize}[leftmargin=9pt,topsep=2pt,itemsep=1pt,parsep=0pt,partopsep=0pt]
\item First, we propose \modelname{}, an asynchronous horizon-adaptive world-action model that decouples slow video-DiT world planning from fast action-DiT closed-loop execution, introducing horizon-adaptive offset training to support arbitrary planner-executor phase relationships.
\item Second, we develop Observation-Guided Video-Context Routing (OVCR), which dynamically constructs chunk-specific latent video context from current observations to keep asynchronous planner context aligned with real-time execution state.
\item Third, we validate \modelname{} across simulation, real-world manipulation, and latency benchmarks, demonstrating state-of-the-art-level manipulation performance without large-scale robot-data pretraining and substantially improved inference efficiency over existing world-action model baselines.
\end{itemize}

\section{Related Work}

\noindent \textbf{Generalist robot manipulation policies.}
Generalist robot policies aim to learn broadly applicable manipulation skills from large-scale diverse demonstrations~\cite{brohan2023can,black2024pi_0,yang2026hivla,liang2024skilldiffuser}. RT-1~\citep{brohan2023rt} and \mbox{RT-2}~\citep{zitkovich2023rt} established the vision-language-action (VLA) paradigm by absorbing heterogeneous robot trajectories and transferring web-scale vision-language knowledge into robotic control. Subsequent works~\citep{kim2025openvla,zheng2025x,pi05,liu2025rdt,openvlaoft,bu2025univla,liang2025discrete,liu2025hybridvla,chen2025unified} further improved generalization by integrating flow-matching and diffusion-based action heads~\citep{chi2025diffusion,janner2022planning} with large pretrained backbones, establishing Diffusion Transformers (DiTs)~\cite{peebles2023scalable} as expressive and scalable policy architectures. However, these methods remain primarily action-centric, with physical scene dynamics only implicitly captured through demonstrations, which motivates the world-action modeling paradigm that \modelname{} extends.

\noindent \textbf{World models for robot control.}
World models~\citep{dreamdojo,hafner2023dreamerv3,wan,seedance} augment robot policies with predictive visual dynamics beyond action-only imitation. Existing video-based robot world models can be broadly grouped into two lines. The first follows an \emph{imagine-then-act} paradigm: methods such as UniPi~\citep{du2023learning}, Seer~\citep{tian2025predictive}, and Video Prediction Policy~\citep{hu2025video} first predict future visual states or predictive visual representations, and then recover actions through inverse dynamics or a policy conditioned on the predicted future, introducing latency in closed-loop deployment. The second line performs \emph{joint world-action modeling}, where future visual dynamics and actions are modeled within a shared generative architecture~\citep{ye2026world, bi2025motus, kim2026cosmos, lingbotva2026, yuan2026fastwam, ye2026gigaworld}. But all couple world prediction and action execution at the same temporal resolution, spending capacity on short-horizon adjacent frames that are redundant for control. \modelname{} addresses this by decoupling the video DiT as a low-frequency long-horizon planner from the action DiT as a high-frequency closed-loop executor.

\noindent \textbf{Dual-system and asynchronous robot policies.}
Recent robot policies have explored dual-system or asynchronous designs to combine deliberative computation with reactive control. RoboDual~\citep{bu2024robodual} uses a VLA generalist to provide task understanding and coarse action guidance for an efficient diffusion specialist, while Reactive Diffusion Policy~\citep{xue2025reactive} combines a slow latent diffusion policy with a fast tactile feedback pathway for contact-rich manipulation. AsyncVLA~\citep{hirose2026asyncvla} runs a large foundation model asynchronously to provide delayed semantic guidance, which is refined by a lightweight edge adapter for fast execution. While these methods validate the effectiveness of dual-system execution, their slow branch primarily provides guidance, feedback, or representations for the action policy. \modelname{} instead realizes the slow-fast split within joint world-action modeling: the video DiT and action DiT operate asynchronously yet remain coupled through layerwise joint attention.

\section{Asynchronous Horizon-Adaptive World-Action Modeling}
\label{sec:method}

\begin{figure}[t]
    \centering
    \includegraphics[width=0.93\linewidth]{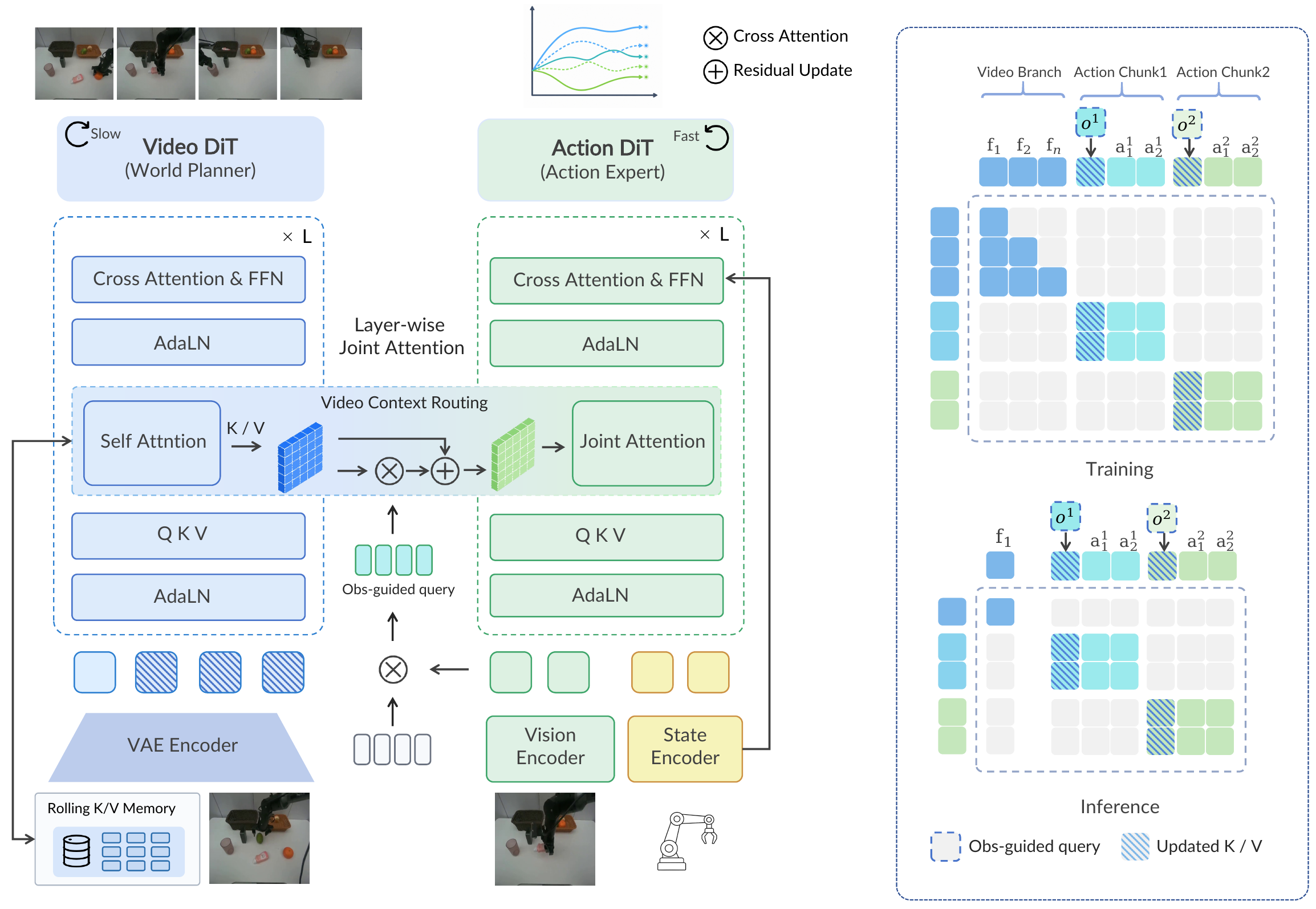}
    \vspace{-2pt}
    \caption{
    \textbf{\modelname{} architecture and attention mask.}
    \modelname{} decouples world planning and action execution into a slow video-DiT planner and a fast action-DiT executor. The video branch is trained with a fully causal mask to learn temporal dynamics. For each action update, the latest observation $o^n$ first queries and updates the video DiT's K/V states through OVCR, producing updated planner context that is consumed by the action DiT through layerwise joint attention.
    }
    \label{fig:model_arch}
    \vspace{-8pt}
\end{figure}

World-action models couple action prediction with learned visual dynamics, but existing formulations typically organize the video branch and the action branch under the same short-horizon execution rhythm. We instead formulate world-action modeling as a two-timescale generation problem. Figure~\ref{fig:model_arch} illustrates the overall architecture.
\modelname{} reorganizes WAM inference into an asynchronous world-action coupling framework while preserving a dual-DiT architecture: a low-frequency video planner produces reusable long-horizon planner context, and a high-frequency action expert consumes this context for closed-loop action denoising. Here, ``horizon-adaptive'' denotes robustness to arbitrary planner--executor phase offsets under decoupled temporal horizons, rather than online horizon selection. In other words, AHA-WAM does not dynamically choose the video or action horizon at test time; instead, it trains the executor to consume long-horizon planner context under variable action-start offsets, making the asynchronous interface robust to the phase misalignment induced by streaming deployment.

This design is supported by three complementary mechanisms.
Observation-Guided Video-Context Routing (OVCR) adapts the cached planner context to the latest observation without rerunning the video DiT.
Horizon-adaptive offset training exposes the action expert to the planner--executor phase shifts induced by asynchronous streaming, while rolling K/V memory extends the video planner's temporal receptive field over past observations.
Finally, real-time inference optimizations further accelerate the high-frequency action stream for closed-loop deployment.

\subsection{Dual-DiT Planner--Executor Architecture}
\label{sec:model_arch}

\noindent \textbf{Problem setting.}
We consider language-conditioned visuomotor policy learning from visual observations and proprioceptive robot states.
At control time $t$, the policy receives a visual observation history $O^v_t$, a proprioceptive state $s_t$, and a language instruction $l$, and predicts an executable action chunk
$A_t = \{a_t,\ldots,a_{t+h_a-1}\}$ of horizon $h_a$.
A standard action-centric policy directly models
\begin{equation}
\pi_\theta(A_t \mid O^v_{\le t}, s_t, l).
\end{equation}

World-action models augment this policy with visual dynamics learning by jointly modeling future visual evolution and action generation.
Let $Z^v_{t:t+h_v}$ denote future video latents over a visual planning horizon $h_v$.
Rather than updating video modeling and action generation under the same short horizon, \modelname{} decouples their temporal roles: the video branch models a longer horizon $h_v$, while the action branch predicts executable chunks of horizon $h_a$, with $h_a<h_v$.

\noindent \textbf{Model architecture.}
\modelname{} is built on a dual-DiT architecture consisting of a video DiT world planner and an action DiT executor.
Visual observations are encoded by the pretrained VAE, while language embeddings from the text encoder condition both branches.
The action branch is a lightweight action expert DiT with the same layer depth as the video DiT, enabling layerwise interaction between the two branches.

The video DiT takes visual latent tokens as input and is trained to predict future video latents over the longer planning horizon.
The action DiT receives noisy action tokens and proprioceptive tokens, and denoises the action chunk under closed-loop robot-state feedback.
Visual feedback for the high-frequency action branch is not introduced by naively concatenating dense visual tokens; instead, visual information is mediated through the planner video context and later adapted by OVCR in Section~\ref{sec:ovcr}.

\noindent \textbf{Layerwise planner--executor coupling.}
The central interface between the two DiTs is the layerwise planner video context.
Given the visual observation context and language instruction, the video planner produces
\begin{equation}
    \mathcal{C}^{p}_{\tau}
    =
    \left\{
    \left(K^{p,\ell}_{\tau}, V^{p,\ell}_{\tau}\right)
    \right\}_{\ell=1}^{L},
    \label{eq:planner_context}
\end{equation}
where $p$ denotes the planner branch, $\tau$ indexes the latest planner refresh and $\ell$ indexes the transformer layer.
This context is a latent world-plan representation exposed by one video-DiT forward and reused by multiple subsequent action-DiT forwards.
It differs from the rolling K/V memory in Section~\ref{sec:rolling_memory}, which stores historical video states across planner refreshes inside the video planner.

During training, the video branch predicts future video latents under a fully causal video mask, encouraging the video DiT to learn forward scene dynamics while shaping the planner context with visual dynamics supervision.
The action branch is masked from attending to future video tokens so that during inference, the future-video prediction path can be removed.
Thus, future-video prediction serves as a world-modeling training signal, while planner video context serves as the inference-time interface to the action executor.

At each action update, the raw planner context $\mathcal{C}^{p}_{\tau}$ is first adapted by OVCR into a chunk-specific context
$\widetilde{\mathcal{C}}^{p}_{t}
=
\{(\widetilde{K}^{p,\ell}_{t},\widetilde{V}^{p,\ell}_{t})\}_{\ell=1}^{L}$.
The action DiT then denoises the current action chunk through layerwise joint attention:
\begin{equation}
    \bar{H}^{a,\ell}_{t}
    =
    \mathrm{Attn}
    \left(
        Q^{a,\ell}_{t},
        \left[K^{a,\ell}_{t}; \widetilde{K}^{p,\ell}_{t}\right],
        \left[V^{a,\ell}_{t}; \widetilde{V}^{p,\ell}_{t}\right]
    \right),
    \label{eq:planner_executor_attention}
\end{equation}
where $Q^{a,\ell}_{t}$, $K^{a,\ell}_{t}$, and $V^{a,\ell}_{t}$ are the action-DiT query, key, and value projections, and $\bar{H}^{a,\ell}_{t}$ is the planner-conditioned action hidden state.
This coupling preserves WAM-style interaction between visual dynamics and action generation, while amortizing expensive video-DiT computation across multiple high-frequency action updates.

\noindent \textbf{Joint world-action training objective.}
\modelname{} is trained with a joint flow-matching objective over action chunks and future video latents.
For a target variable $y$, either an action chunk $A_t$ or future video latents $Z^v_{t:t+h_v}$, we sample Gaussian noise $\epsilon \sim \mathcal{N}(0,I)$ and a flow time $\rho \in (0,1)$, and construct the interpolated sample
\begin{equation}
    y_{\rho}
    =
    (1-\rho)y + \rho \epsilon .
\end{equation}
The model predicts the corresponding velocity field:
\begin{equation}
    \mathcal{L}_{\mathrm{FM}}(y)
    =
    \mathbb{E}_{y,\epsilon,\rho}
    \left[
    \left\|
    f_\theta(y_{\rho}, \rho, O^v_t, s_t, l)
    -
    (\epsilon-y)
    \right\|_2^2
    \right].
    \label{eq:fm_loss}
\end{equation}
We instantiate this objective for action generation and video co-training:
\begin{equation}
    \mathcal{L}_{a}
    =
    \mathcal{L}_{\mathrm{FM}}(A_t),
    \qquad
    \mathcal{L}_{v}
    =
    \mathcal{L}_{\mathrm{FM}}(Z^v_{t:t+h_v}).
\end{equation}
The final objective is
\begin{equation}
    \mathcal{L}
    =
    \mathcal{L}_{a}
    +
    \lambda \mathcal{L}_{v},
    \label{eq:joint_objective}
\end{equation}
where $\lambda$ balances action learning and video dynamics learning.
At deployment, \modelname{} removes explicit future-frame decoding: the video DiT only refreshes planner video context, and the action DiT uses this context to generate closed-loop action chunks.

\subsection{Observation-Guided Video-Context Routing}
\label{sec:ovcr}

Asynchronous execution reuses one planner video context for multiple action chunks, which amortizes video-DiT computation but creates a context-alignment problem: before the next planner refresh, the robot state and visual scene may have changed.
Observation-Guided Video-Context Routing (OVCR) addresses this by using the latest visual observation to convert the shared planner context into a chunk-specific context.
Instead of feeding dense visual tokens into the high-frequency action DiT, OVCR compresses them into a small set of routing queries that select and edit the relevant planner context before action denoising.

For each action chunk at time $t$, we split the aligned observation context into visual tokens $X^v_t$ and proprioceptive tokens $X^s_t$.
The proprioceptive input is compact and directly tied to the instantaneous robot state, so a lightweight encoder maps it to a state token which is directly provided to the action DiT.
Visual feedback is injected indirectly through context routing.
Given $Q$ learnable base queries $B \in \mathbb{R}^{Q \times d}$, OVCR constructs observation-guided routing queries by attention pooling over the current visual tokens:
\begin{equation}
    Z^q_t
    =
    \mathrm{Attn}
    \left(
        B,
        f_v(X^v_t),
        f_v(X^v_t)
    \right),
    \label{eq:ovcr_query}
\end{equation}
where $f_v$ is a lightweight visual projection module.
The resulting queries provide compact, observation-conditioned slots for retrieving chunk-relevant information from the planner video context.

Let
$\mathcal{C}^{p}_{\tau(t)}
=
\{(K^{p,\ell}_{\tau(t)}, V^{p,\ell}_{\tau(t)})\}_{\ell=1}^{L}$
denote the latest available planner context.
For each layer $\ell$, OVCR first reads planner features using the routing queries and then predicts residual key--value updates:
\begin{equation}
    R^{\ell}_t
    =
    \mathrm{Attn}
    \left(
        Z^{q,\ell}_t,
        K^{p,\ell}_{\tau(t)},
        V^{p,\ell}_{\tau(t)}
    \right),
    \qquad
    (\Delta K^{p,\ell}_t, \Delta V^{p,\ell}_t)
    =
    g^{\ell}_{\psi}
    \left(
        R^{\ell}_t,
        Z^{q,\ell}_t
    \right),
    \label{eq:ovcr_delta}
\end{equation}
where $g^{\ell}_{\psi}$ is a lightweight layerwise router.
The chunk-specific planner context is produced by a gated residual update:
\begin{equation}
    \widetilde{K}^{p,\ell}_t
    =
    K^{p,\ell}_{\tau(t)}
    +
    \alpha^{\ell}_t \Delta K^{p,\ell}_t,
    \qquad
    \widetilde{V}^{p,\ell}_t
    =
    V^{p,\ell}_{\tau(t)}
    +
    \alpha^{\ell}_t \Delta V^{p,\ell}_t,
    \label{eq:ovcr_update}
\end{equation}
where $\alpha^{\ell}_t$ is a learned gate.
The adapted context
$\widetilde{\mathcal{C}}^{p}_{t}
=
\{(\widetilde{K}^{p,\ell}_t,\widetilde{V}^{p,\ell}_t)\}_{\ell=1}^{L}$
is then consumed by the action DiT through the layerwise joint attention in Eq.~\ref{eq:planner_executor_attention}.

OVCR turns planner-context reuse from static caching into observation-conditioned retrieval and adaptation.
Thus, the video DiT remains outside the per-update critical path, while each action chunk still receives a planner representation aligned with the latest visual evidence.

\subsection{Horizon-Adaptive Offset Training}
\label{sec:offset_training}

\begin{figure}[t]
    \centering
    \includegraphics[width=0.99\linewidth]{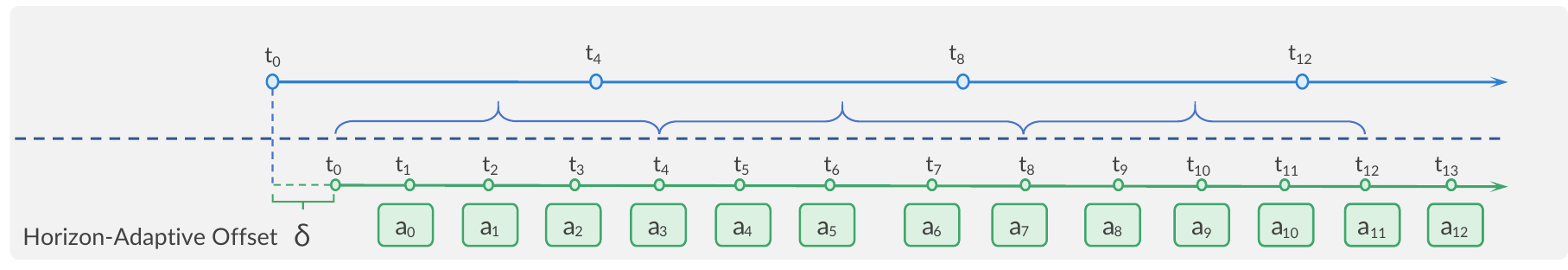}
    \vspace{-4pt}
    \caption{
    \textbf{Horizon-adaptive offset training.}
    We randomly shift the action-chunk grid by $\delta \in [0,h_a)$ inside the video planning horizon, so the action executor learns to consume planner context under different phase offsets induced by asynchronous deployment.
    }
    \label{fig:asy}
    \vspace{-8pt}
\end{figure}

Asynchronous streaming changes the relative temporal phase between the slow planner and the fast executor.
If training always uses a fixed alignment between the video planning window and the action chunk, the action DiT may overfit to a single planner--executor phase and become brittle when reusing planner context at intermediate phases during deployment.
We therefore introduce horizon-adaptive offset training to expose the executor to the phase shifts induced by asynchronous inference.

Let $h_v$ denote the video planning horizon and $h_a$ denote the action chunk horizon, with $h_a<h_v$.
As shown in Figure~\ref{fig:asy}, for each training segment starting at time $\tau$, the video planner models future video latents over $[\tau,\tau+h_v)$.
Instead of aligning the first action chunk to the planner start, we sample
\begin{equation}
    \delta \sim \mathcal{U}\{0,1,\ldots,h_a-1\},
    \label{eq:offset_sample}
\end{equation}
and shift the action-chunk grid by $\delta$ within the planner horizon.
The action objective is then evaluated over offset-aligned chunks:
\begin{equation}
    \mathcal{L}_{a}
    =
    \mathbb{E}_{\delta}
    \left[
    \mathcal{L}_{\mathrm{FM}}
    \left(
    A^{\delta}_{\tau}
    \right)
    \right],
    \label{eq:offset_action_loss}
\end{equation}
where $A^{\delta}_{\tau}$ denotes an action chunk starting from an offset-aligned position inside the planner horizon, and $\mathcal{L}_{\mathrm{FM}}$ is defined in Eq.~\ref{eq:fm_loss}.

Since the planner--executor alignment is periodic with the action chunk size, sampling $\delta \in [0,h_a)$ covers all chunk-level phases encountered when one planner context is reused across multiple executor updates.
Thus, horizon-adaptive offset training teaches the action DiT to consume long-horizon planner context under variable action-start offsets, while OVCR handles observation-conditioned context selection at each phase.

\subsection{Rolling Planner Memory}
\label{sec:rolling_memory}

Since the video DiT operates as a low-frequency planner, it should preserve historical scene information across planner refreshes rather than relying only on the current observation.
This is important for long-horizon manipulation, where completed subgoals, displaced objects, or previously observed states can provide essential context.
We therefore maintain a fixed-size FIFO rolling K/V memory inside the video planner.

For each layer $\ell$, the memory stores historical video states from recent planner refreshes:
\begin{equation}
    M^\ell_\tau =
\mathrm{FIFO}
\left(
M^\ell_{\tau-1}
\cup
\{(K^{p,\ell}_\tau,V^{p,\ell}_\tau)\}
\right),
    \label{eq:rolling_memory_update}
\end{equation}
where $\tau$ indexes planner refreshes and the memory window size is fixed.
At the next refresh, the video DiT attends to this memory when producing the new planner video context
$\mathcal{C}^{p}_{\tau}
=
\{(K^{p,\ell}_{\tau}, V^{p,\ell}_{\tau})\}_{\ell=1}^{L}$.

This memory is internal to the slow video planner and is not directly consumed by the action DiT.
It extends the planner's temporal receptive field before the next planner context is produced, while the high-frequency executor still interacts only with the latest OVCR-adapted planner video context.

\subsection{Streaming Inference and Real-Time Optimization}
\label{sec:inference}

The asynchronous planner-executor architecture naturally leads to a streaming inference schedule.
At deployment, the video planner and the action executor are executed as two non-blocking streams, where \modelname{} removes the video DiT from the per-update critical path, but the action branch must still run at the closed-loop control frequency. We therefore optimize the action-chunk inference path, measuring the end-to-end latency $L_{\text{chunk}}$ of one action update, including image encoding, planner-context access, OVCR routing, and action denoising. Video-DiT prefill is executed asynchronously, so $L_{\text{chunk}}$ directly determines the action-update frequency reported in Section~\ref{sec:latency}.

\noindent \textbf{CUDA acceleration.}
We compile the repeated action-phase computation into a static deployment path by executing the action DiT, memory and context modules, and VAE encoder through TensorRT and CUDA-graph capture where possible, while removing redundant loop-invariant computations and repeated buffer copies from the denoising hot path. These changes do not alter the model architecture, weights, or sampling procedure, but reduce the 10-step action inference latency from $415.77$ ms in PyTorch eager execution to $41.37$ ms.

\noindent \textbf{ODE distillation.}
On top of the CUDA-accelerated 10-step path, we construct \modelname{}-Flash by distilling the action sampler from 10 denoising steps to 2 steps while keeping the same observation, planner-context, and OVCR interface. The video DiT is frozen during distillation, and the student is trained from sampled intermediate states to directly predict the teacher's final action output, with sampling biased toward noisier states to support aggressive step reduction. ODE distillation further reduces $L_{\text{chunk}}$ from $41.37$ ms to $17.56$ ms. Additional CUDA ablations, distillation schedules, and step-latency tradeoffs are provided in Appendix~\ref{app:inference_speedup}.

\section{Experiments}
\label{sec:experiments}

We evaluate \modelname{} in simulation and on real robots to test whether asynchronous world-action modeling preserves manipulation performance while improving closed-loop efficiency.
We cover RoboTwin 2.0~\citep{chen2025robotwin} performance, component ablations, real-world deployment, generalization analysis and inference latency.

Specifically, our experiments ask five questions including capability, mechanism, deployability, robustness, and efficiency.
\textbf{Q1.} Can our asynchronous WAM preserve or improve simulation performance while faster? (Sec. ~\ref{sec:robotwin})
\textbf{Q2.} How important is each component for \modelname{}? (Sec. ~\ref{sec:ablation})
\textbf{Q3.}  Can \modelname{} achieve reliable real-robot performance? (Sec.~\ref{sec:real_world})
\textbf{Q4.} How robust is \modelname{} under previously unseen real-world task conditions? (Sec.~\ref{sec:real_world})
\textbf{Q5.} How much closed-loop latency reduction does \modelname{} provide? (Sec.~\ref{sec:latency})

\subsection{Experimental Setup}
\label{sec:exp_setup}

\noindent\textbf{Model and training.}
\modelname{} is implemented as a dual-DiT policy with an explicitly asynchronous planner--executor interface.
We use the pretrained Wan2.2-5B video model to initialize the world-planning branch, including the video DiT, text encoder, and video VAE.
Following Fast-WAM~\citep{yuan2026fastwam}, the high-frequency action executor adopts a compact DiT architecture with hidden dimension $d_a=1024$, corresponding to approximately $1.02$B parameters.
In addition to the $4.99$B-parameter video planner and the $1.02$B-parameter action DiT, \modelname{} introduces $1.22$B parameters for rolling K/V memory and context-routing modules, giving a total instantiated model size of about $7.23$B parameters.

For the temporal configuration of \modelname{}, the video branch operates over a long planning horizon of $h_v=64$, while the action branch predicts short executable chunks with horizon $h_a=16$.
During asynchronous inference, the video planner maintains reusable layerwise K/V context, which is augmented by a FIFO rolling memory over at most 6 historical observation frames.
The action executor adapts this context through OVCR, using 32 observation-guided routing queries for each action chunk.

Training uses flow matching for both world modeling and action prediction, with logit-normal sampling of noise times.
For action generation, we use 10 denoising steps at inference and set CFG to 1.0.
All models are trained with AdamW, a learning rate of $1\times10^{-4}$, weight decay $0.01$, cosine scheduling, mixed precision, and gradient clipping.
Latency is reported on a single NVIDIA RTX 5090D GPU.
Appendix~\ref{app:config} lists additional implementation details, including \modelname{}-Flash distillation and other training configurations.

\par\vspace{-2pt}

\noindent\textbf{Evaluation scope.}
We evaluate \modelname{} in both RoboTwin 2.0 simulation~\citep{chen2025robotwin} and real-world experiment. The simulation benchmark tests multi-task manipulation under clean and randomized scenes, while the real-world experiments test deployment on physical bimanual tasks. Across both settings, we report task success rate as the primary metric and additionally analyze closed-loop inference latency to quantify deployment efficiency.

\subsection{RoboTwin Simulation}
\label{sec:robotwin}
We evaluate \modelname{} on RoboTwin 2.0~\citep{chen2025robotwin}, a bimanual manipulation benchmark with 50 dual-arm tasks covering diverse skills.
We follow the multi-task training setup of~\citep{yuan2026fastwam, bi2025motus, lingbotva2026}: for each task, the training set contains 50 demonstrations in the clean setting and 500 demonstrations in the randomized setting.
Models are trained in a multi-task manner with a global batch size of 512.
Each method is tested over 100 trials per task in both the clean and randomized settings, and we report task-averaged success rates.
We compare against competitive VLA and WAM baselines, including $\pi_0$~\citep{black2024pi_0}, $\pi_{0.5}$~\citep{pi05}, ABot-M0~\citep{yang2026abot}, Motus~\citep{bi2025motus}, LingBot-VA~\citep{lingbotva2026}, and Fast-WAM~\citep{yuan2026fastwam}.
Appendices~\ref{app:robotwin_details} and~\ref{app:per_task_success} provide detailed protocols, baseline settings, and per-task success rates.

\begin{table}[t]
    \centering
    \setlength{\abovecaptionskip}{1pt}
    \setlength{\belowcaptionskip}{1pt}
    \caption{\textbf{RoboTwin 2.0 average success on $50$ tasks.}}
    \vspace{2pt}
    \label{tab:robotwin_main}
    \footnotesize
    \renewcommand{\arraystretch}{0.92}
    \resizebox{0.74\linewidth}{!}{
    \begin{tabular}{lcccc}
        \toprule
        Method & Robo. P.T. & Clean (\%) & Rand. (\%) & Avg. (\%) \\
        \midrule
        $\pi_0$ & \cmark & 65.92 & 58.40 & 62.16 \\
        $\pi_{0.5}$ & \cmark & 82.74 & 76.76 & 79.75 \\
        ABot-M0 & \cmark & 81.20 & 80.40 & 80.80 \\
        Motus from Wan2.2 & \xmark & 77.56 & 77.00 & 77.28 \\
        Motus & \cmark & 88.66 & 87.02 & 87.84 \\
        LingBot-VA & \cmark & \cellcolor{secondgreen}92.90 & \cellcolor{secondgreen}91.50 & \cellcolor{secondgreen}92.20 \\
        Fast-WAM & \xmark & 91.88 & 91.78 & 91.83 \\
        \midrule
        \modelname{}-Flash & \xmark & 90.48 & 89.92 & 90.20 \\
        \modelname{} & \xmark & \cellcolor{bestgreen}\textbf{93.40} & \cellcolor{bestgreen}\textbf{92.20} & \cellcolor{bestgreen}\textbf{92.80} \\
        \bottomrule
    \end{tabular}}
    \vspace{-8pt}
\end{table}

\noindent \textbf{Results and analysis.}
Table~\ref{tab:robotwin_main} summarizes the RoboTwin results: \modelname{} achieves $93.40\%$ success in the clean setting and $92.20\%$ under randomized evaluation, yielding an average success rate of $92.80\%$.
Without robot data pretraining, \modelname{} improves over Fast-WAM by $0.97$ percentage points, showing that our asynchronous design not only increases the closed-loop control frequency but also preserves and improves performance.
\modelname{} also exceeds LingBot-VA, the strongest baseline with large-scale robot-data pretraining, by $0.60$ points.
The \modelname{}-Flash variant retains $90.20\%$ average success, incurring only a modest performance drop while further cutting the latency of \modelname{} by more than half as analyzed in Section~\ref{sec:latency}.

\FloatBarrier

\subsection{Ablation Studies on RoboTwin}
\label{sec:ablation}

\begin{wraptable}{r}{0.5\linewidth}
    \vspace{-12pt}
    \centering
    \setlength{\abovecaptionskip}{2pt}
    \setlength{\belowcaptionskip}{2pt}
    \setlength{\tabcolsep}{3pt}
    \scriptsize
    \caption{
    \textbf{RoboTwin 2.0 ablations.}
    }
    \label{tab:robotwin_ablation}
    \resizebox{\linewidth}{!}{
    \begin{tabular}{lccc}
        \toprule
        Variant & Clean (\%) & Rand. (\%) & Avg. (\%) \\
        \midrule
        Fast-WAM & 91.88 & 91.78 & 91.83 \\
        Naive-Async & 88.64 & 88.56 & 88.60 \\
        + KV Memory & 91.40 & 90.62 & 91.01 \\
        + OVCR & 91.52 & 91.42 & 91.47 \\
        \modelname{} & \textbf{93.40} & \textbf{92.20} & \textbf{92.80} \\
        \bottomrule
    \end{tabular}}
    \vspace{-8pt}
\end{wraptable}

We ablate the components that make asynchronous world-action modeling effective.
Naive-Async decouples the video and action branches, but directly reuses the latest planner context without rolling K/V memory or OVCR.
We then add each mechanism separately and jointly.

Table~\ref{tab:robotwin_ablation} shows that asynchronous execution alone is insufficient.
Naive-Async drops from $91.83\%$ to $88.60\%$, indicating that faster action updates cannot compensate for stale or phase-misaligned planner context.
Adding rolling K/V memory recovers performance to $91.01\%$, showing that persistent planner states help stabilize the reused video context across refreshes.
The gain is moderate, which is expected because most RoboTwin tasks are short-to-medium horizon and keep task-relevant objects visible.
OVCR further improves success to $91.47\%$, suggesting that observation-conditioned context adaptation is the more direct remedy for asynchronous planner--executor mismatch.
The full \modelname{} reaches $92.80\%$, confirming that memory and routing are complementary: memory preserves temporal context, while OVCR aligns it with the current execution state.

\subsection{Real-World Robot Experiments}
\label{sec:real_world}

\noindent\textbf{Setup.}
We evaluate on a bimanual AgileX Piper platform with an ego-view RGB camera, and policies uses only head-view RGB observations, proprioceptive states and language instructions as input.
We consider four tasks---\textit{Fold Towel}, \textit{Organize Desktop}, \textit{Prepare Soy Milk}, and \textit{Store Plate}---covering deformable manipulation, long-horizon rearrangement, fine-grained tool use, contact-rich control, and spatial generalization. For each task, we collect approximately $120$ episodes on average.
Appendix~\ref{app:task_execution_illustration} provides task details, data scale, execution procedures, and generalization variants.
\par\vspace{-2pt}

\noindent\textbf{Implementation.}
Since Fast-WAM and AHA-WAM are not pretrained on any robot data by default, we pretrain Fast-WAM and \modelname{} on the selected RoboCOIN subset~\citep{RoboCOINReport}, containing $24{,}600$ trajectories and approximately $165$ hours of robot data for a fair and stable deployment comparison.
Both models are finetuned on the same task-specific demonstrations.
Motus and Fast-WAM incur large naive deployment latency, which leads to sparse action updates and unstable closed-loop behavior.
We deploy them with an RTC-style non-blocking execution scheme~\citep{black2026real} and action interpolate between predicted chunks, enabling smoother control under high inference latency.
\modelname{} is deployed with its asynchronous planner--executor interface, where the executor performs high-frequency closed-loop updates using the latest observation and reused planner context.
\par\vspace{-2pt}

\begin{figure}[t]
    \centering
    \includegraphics[width=0.95\linewidth]{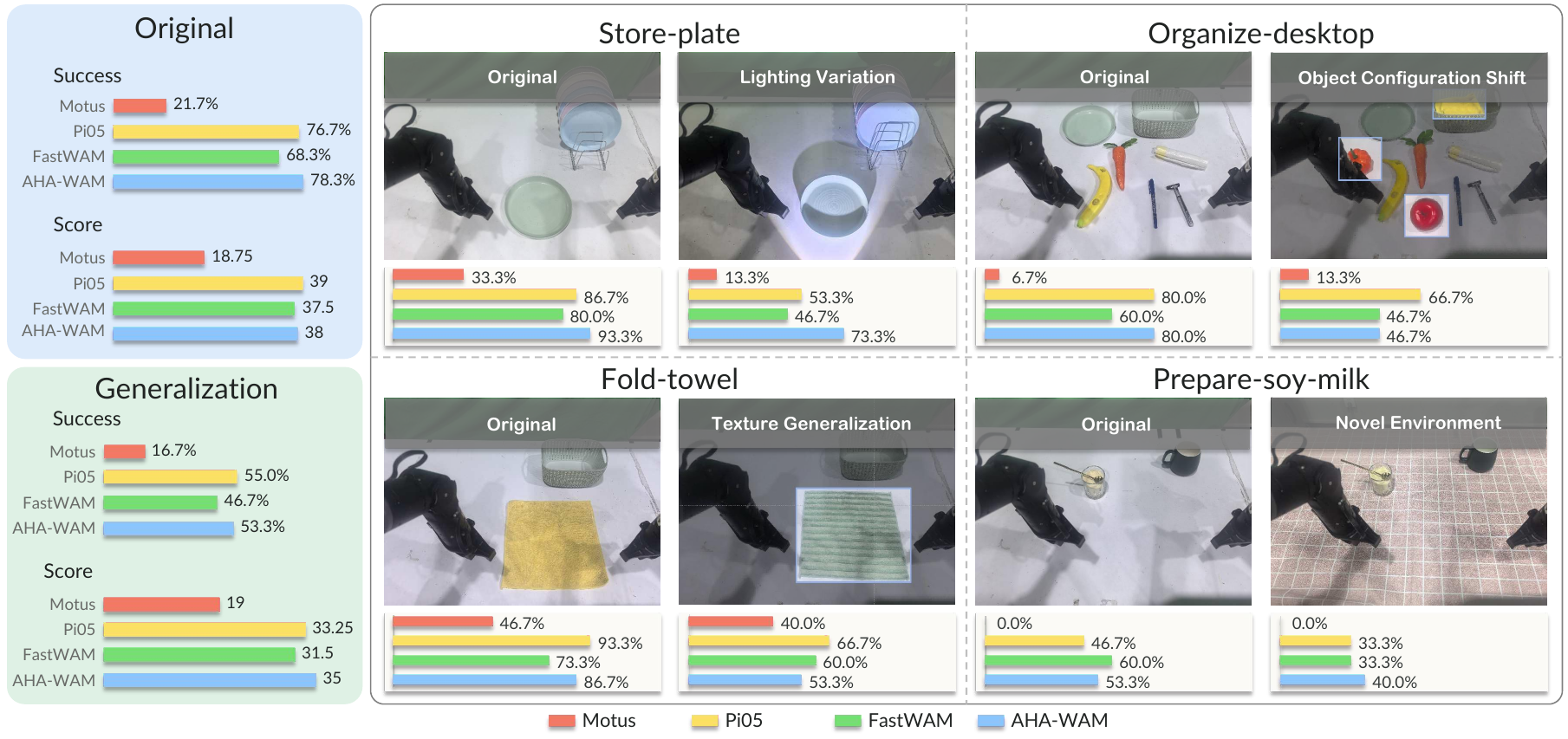}
    \vspace{-8pt}
    \caption{\textbf{Real-world task success rates and scores.} Success and 0--3 task scores are computed over 30 trials; scoring criteria are in Appendix~\ref{app:task_execution_illustration}.}
    \label{fig:real_world_exp}
    \vspace{-4pt}
\end{figure}

\noindent\textbf{Results.}
Figure~\ref{fig:real_world_exp} reports real-world performance under original settings and generalization shifts.
In the original settings, \modelname{} achieves $78.33\%$ success, clearly outperforming the WAM baselines Motus ($21.67\%$) and Fast-WAM ($68.33\%$), while matching the strong generalist VLA baseline $\pi_{0.5}$ ($76.67\%$).
Under generalization shifts, $\pi_{0.5}$ obtains the highest success rate, while \modelname{} remains second in success and achieves the highest progress score ($35.00$).
This suggests that, despite not relying on $\pi_{0.5}$-scale generalist pretraining, \modelname{} attains comparable real-world robustness and delivers the strongest deployment performance among WAM-based baselines.
These results indicate that asynchronous world-action modeling improves deployment stability: long-horizon planner context supports task progress, while high-frequency execution enables timely physical correction.

\subsection{Latency and Control Frequency}
\label{sec:latency}

\begin{wraptable}[5]{r}{0.5\linewidth}
    \vspace{-20pt}
    \centering
    \setlength{\abovecaptionskip}{1pt}
    \setlength{\belowcaptionskip}{1pt}
    \setlength{\tabcolsep}{1.5pt}

    \caption{
    Inference latency and frequency.
    }
    \label{tab:latency}
    \resizebox{\linewidth}{!}{
    \begin{tabular}{lccc}
        \toprule
        Method & Lat. (ms) & Freq. (Hz) & Speedup \\
        \midrule
        Motus & 1866.10 & 0.54 & $0.10\times$ \\
        Fast-WAM & 190.00 & 5.26 & $1.00\times$ \\
        \modelname{} & 41.37 & 24.17 & $4.59\times$ \\
        \modelname{}-Flash & \textbf{17.56} & \textbf{56.95} & \textbf{$10.82\times$} \\
        \bottomrule
    \end{tabular}
    }
    \vspace{-5pt}
\end{wraptable}

Finally, we compare closed-loop inference latency in
\mbox{Tab.~\ref{tab:latency}}. Fast-WAM is reported with official latency, while the remainings are measured on a single NVIDIA RTX 5090D GPU.
Appendix~\ref{app:inference_speedup} gives protocol and optimization details.

Compared with Fast-WAM, \modelname{} reduces latency from $190.00$ ms to $41.37$ ms, improving the closed-loop rate from $5.26$ Hz to $24.17$ Hz.
With the distilled sampler, \modelname{}-Flash further reaches $17.56$ ms and $56.95$ Hz, yielding a $10.82\times$ speedup while keeping the same asynchronous planner--executor interface.

\section{Conclusion}
\label{sec:conclusion}

We presented \modelname{}, an asynchronous horizon-adaptive world-action model that decouples a low-frequency video DiT world planner from a high-frequency action DiT closed-loop executor. Observation-Guided Video-Context Routing and horizon-adaptive offset training make this asynchronous interface effective by adapting reused planner context to each current observation and handling arbitrary planner-executor phase offsets. Experiments on RoboTwin and real-world manipulation tasks demonstrate strong performance and substantially reduced closed-loop latency, showing that learned visual dynamics can improve robot control without compromising control frequency.

\paragraph{Limitations and future work.}
The planner update frequency, video horizon, and action chunk size introduce temporal hyperparameters whose optimal allocation may depend on task dynamics and embodiment. Future work could strengthen the slow planner with longer-horizon prediction and richer scene representations, while systematic evaluation on dedicated long-horizon benchmarks would better quantify how the asynchronous design scales with task complexity.

More broadly, \modelname{} opens a new design space for WAMs.
Once the video branch is decoupled from the high-frequency control loop, it can afford more expensive computation without directly increasing action latency. The asynchronous interface makes these extensions particularly attractive: the video branch can become more predictive, deliberative, or physically grounded, while the action branch preserves the fast closed-loop rate needed for deployment.
We believe this separation between slow scalable world planning and fast reactive action execution is a promising direction for building more capable, efficient, and deployable world-action models.

\section*{Acknowledgments}

We would like to express our sincere gratitude to the Baidu AI Cloud Baige Team for their exceptional technical support and for providing access to the state-of-the-art Baidu AIHC platform. We specifically appreciate the platform's powerful capabilities in delivering efficient training acceleration and enabling ultra-low-latency inference, which were instrumental in optimizing our system's performance during evaluation. The advanced system optimizations and robust distributed infrastructure provided by this team were crucial in accelerating our experiments and validating the scalability of our proposed methods.

\clearpage
{
\bibliography{bibliography_custom}
}

\clearpage

\appendix

\section{Implementation Settings}
\label{app:config}

This appendix records implementation details that complement the experimental setup in Section~\ref{sec:exp_setup}.
The main text describes the benchmark-level settings and the asynchronous planner--executor schedule; here we summarize the concrete implementation settings used for the RoboTwin runs.
We omit logging paths, checkpoint paths, and inactive legacy annealing options.

\paragraph{Model architecture.}
Table~\ref{tab:aha_wam_settings} provides the key architecture and hyperparameter settings for \modelname{} on RoboTwin, including the Video-DiT planner, Action-DiT executor, and OVCR interface.
The instantiated model contains approximately $4.99$B parameters in the video branch, $1.02$B parameters in the action DiT, and $1.22$B parameters in the memory and context-routing modules.
During asynchronous inference, the action path receives the latest observation and proprioceptive state while querying the most recently available planner context through OVCR.

\begingroup
\setlength{\tabcolsep}{2.5pt}
\renewcommand{\arraystretch}{0.82}
\begin{table}[!htbp]
    \centering
    \scriptsize
    \caption{Implementation settings for \modelname{} on RoboTwin.}
    \label{tab:aha_wam_settings}
    \begin{tabular}{@{}>{\raggedright\arraybackslash}p{0.23\linewidth}>{\raggedright\arraybackslash}p{0.29\linewidth}>{\raggedright\arraybackslash}p{0.40\linewidth}@{}}
    \toprule
    Component & Subcomponent & Setting \\
    \midrule
    \multicolumn{3}{@{}l}{\textit{Component settings}} \\
    Video-DiT & Backbone & Wan2.2-TI2V-5B video expert \\
     & Observation cameras & $3$ views: head, left wrist, right wrist \\
     & Image resolution & $384\times320$ \\
     & History frames & $6$ \\
     & Video/action frequency ratio & $8$ \\
     & Video RoPE stride & $8$ \\
     & Routed transformer layers & $30$ \\
     & Attention heads / head dimension & $24$ / $128$ \\
    \midrule
    Action-DiT & Action horizon / chunk size & $64$ actions / $16$ actions \\
     & Number of action chunks & $4$ \\
     & State / proprio dimension & $14$ \\
     & Action dimension & $14$ \\
     & Training objective & Action prediction with planner-context conditioning \\
    \midrule
    OVCR & Method name & Observation-Guided Video-Context Routing \\
     & Query source & Visual observation context \\
     & Queries per chunk & $32$ \\
     & Target context & Causal video K/V cache \\
     & Granularity & Per action chunk, per transformer layer \\
    \midrule
    \multicolumn{3}{@{}l}{\textit{Hyperparameter settings}} \\
    Training & Optimizer / learning rate / weight decay & AdamW / $1\times10^{-4}$ / $1\times10^{-2}$ \\
     & Learning-rate schedule / warmup & Cosine / first $5\%$ of training \\
     & Global batch size / epochs & $512$ / $5$ \\
     & Dataloader workers & $16$ \\
     & Video/action train timesteps & $1000$ \\
     & Video/action train and inference shift & $5.0$ \\
    \midrule
    Inference & Default action denoising steps / CFG & $10$ / $1.0$ \\
    \bottomrule
    \end{tabular}
\end{table}
\endgroup

\paragraph{ODE distillation.}
Table~\ref{tab:ode_distillation_settings} lists the hyperparameter settings used for ODE-distilled fast sampling.
It records the distillation target, teacher--student denoising schedules, and the main optimization settings; dataset paths, checkpoint paths, logging paths, and other run-specific bookkeeping are omitted.

\begingroup
\setlength{\tabcolsep}{4pt}
\renewcommand{\arraystretch}{0.95}
\begin{table}[!htbp]
    \centering
    \footnotesize
    \caption{Hyperparameter settings for ODE-distilled fast sampling.}
    \label{tab:ode_distillation_settings}
    \begin{tabular}{@{}>{\raggedright\arraybackslash}p{0.42\linewidth}>{\raggedright\arraybackslash}p{0.50\linewidth}@{}}
    \toprule
    Item & Setting \\
    \midrule
    Distillation target & Action ODE / flow trajectory \\
    Teacher denoising steps & $16$ \\
    Student denoising steps & $2$ \\
    Distilled timesteps & $5{,}000$ \\
    Teacher capture schedule & $0, 1, 2, 4, 8, 12, 16$ \\
    Prediction parameterization & Flow \\
    Global batch size & $512$ \\
    Learning rate & $2\times10^{-5}$ \\
    Optimizer / weight decay & AdamW / $1\times10^{-2}$ \\
    Learning-rate schedule & Cosine \\
    Epochs & $5$ \\
    Local batch size / gradient accumulation & $16$ / $4$ \\
    Dataloader workers & $16$ \\
    \bottomrule
    \end{tabular}
\end{table}
\endgroup

\paragraph{Training configuration.}
The optimizer, scheduler, batch-size, epoch, and default inference settings for \modelname{} are summarized in Table~\ref{tab:aha_wam_settings}, while the corresponding ODE-distillation settings are summarized in Table~\ref{tab:ode_distillation_settings}.
We use flow matching for both video prediction and action prediction, with noise times sampled from a logit-normal distribution; when both branches are optimized jointly, the video loss and action loss are weighted equally.
Horizon-adaptive offset training follows Section~\ref{sec:offset_training}, so each action chunk is trained to consume planner context under randomized planner--executor phase offsets.
The distilled \modelname{}-Flash variant uses the reduced-step sampler described in Appendix~\ref{app:inference_speedup}.

\section{RoboTwin Evaluation Details}
\label{app:robotwin_details}

This appendix provides the benchmark and baseline details omitted from the main RoboTwin discussion in Section~\ref{sec:robotwin}.

\paragraph{Benchmark protocol.}
RoboTwin 2.0 is evaluated with the AgileX embodiment and contains $50$ dual-arm manipulation tasks spanning a broad range of bimanual skills.
Following the multi-task setting used by prior WAM baselines, \modelname{} is trained with $50$ clean demonstrations and $500$ randomized demonstrations per task, totaling $2{,}500$ clean and $25{,}000$ randomized demonstrations.
Each task is evaluated with $100$ clean episodes and $100$ randomized episodes, and we report task-averaged success rates.
The randomized setting introduces visual and scene-level variations, providing a stronger robustness test than clean evaluation.

\paragraph{Baseline comparison.}
We compare \modelname{} with representative VLA and WAM baselines.
Fast-WAM is the closest baseline because it also uses a video DiT inside a world-action model, while Motus and LingBot-VA represent unified world-action modeling architectures.
$\pi_0$, $\pi_{0.5}$, and ABot-M0 provide comparisons to strong generalist VLA policies.
When available, we include both embodied-pretrained and Wan2.2-initialized variants to separate the effect of robot-data pretraining from the model architecture.
In Table~\ref{tab:robotwin_main}, Embodied PT. indicates whether the method uses embodied robot-data pretraining before RoboTwin 2.0 training.
We report official published RoboTwin 2.0 results for external baselines when available.

\section{Real-World Task Execution and Scoring Criteria}
\label{app:task_execution_illustration}

\begin{figure}[t]
    \centering
    \includegraphics[width=0.98\linewidth]{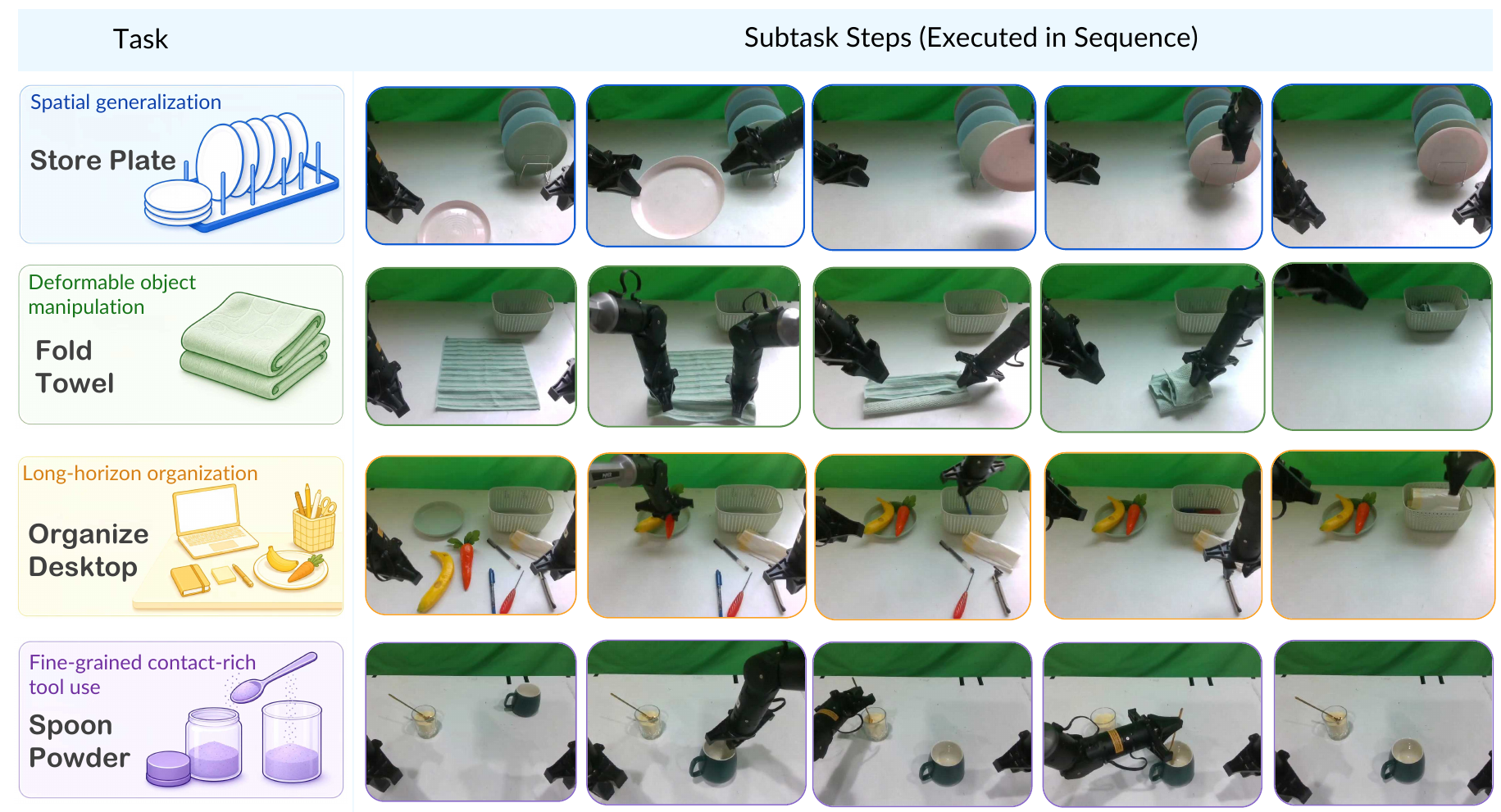}
    \vspace{-6pt}
    \caption{Comprehensive illustration of the real-world task execution process. Each row corresponds to one task type, and the numbered panels show the ordered subtask steps executed in sequence.}
    \label{fig:task_exec_appendix}
    \vspace{-8pt}
\end{figure}

Figure~\ref{fig:task_exec_appendix} provides a step-by-step illustration of the four real-world manipulation tasks used in our evaluation.
Each task is decomposed into three ordered subtask steps, which are executed sequentially during a rollout and directly define the partial-progress score used in Figure~\ref{fig:real_world_exp}.
For each model and task, success rate and score are computed over 30 independent trials.
Success is recorded as a binary outcome, while the score is assigned on a 0--3 scale: a score of 0 indicates no meaningful task progress, and scores of 1, 2, and 3 indicate completion of subtask steps respectively.
The task is counted as success when the rollout reaches score 3.
This unified execution-and-scoring protocol makes the evaluation more informative than a binary success label alone, because it distinguishes early failures from trials that complete most of the required manipulation sequence.

\begin{table}[t]
    \centering
    \scriptsize
    \setlength{\tabcolsep}{2.5pt}
    \renewcommand{\arraystretch}{1.25}
    \caption{Summary of real-world task objectives, partial-progress scoring criteria, and evaluated control challenges. All tasks use a unified 0--3 score, where score 0 indicates no meaningful progress and score 3 indicates full task completion.}
    \label{tab:real_world_scoring}
    \begin{tabular}{@{}>{\raggedright\arraybackslash}p{0.12\linewidth}>{\raggedright\arraybackslash}p{0.17\linewidth}>{\centering\arraybackslash}p{0.07\linewidth}>{\raggedright\arraybackslash}p{0.36\linewidth}>{\raggedright\arraybackslash}p{0.21\linewidth}@{}}
        \toprule
        Task & Objective & Score & Completion criterion & Evaluated challenges \\
        \midrule
        \multirow{3}{=}{Store Plate} &
        \multirow{3}{=}{Move a plate into a plate rack.} &
        1 & The left hand securely grasps the plate. &
        \multirow{3}{=}{Bimanual transfer, spatial placement, and pose control.} \\
        & & 2 & The plate is handed over from the left hand to the right hand. & \\
        & & 3 & The right hand places the plate into the rack. & \\
        \midrule
        \multirow{3}{=}{Fold Towel} &
        \multirow{3}{=}{Fold a towel and place it into a basket.} &
        1 & Both hands fold the towel forward. &
        \multirow{3}{=}{Deformable-object handling and robustness to shape changes.} \\
        & & 2 & The right hand folds the towel once more. & \\
        & & 3 & The right hand deposits the folded towel into the basket. & \\
        \midrule
        \multirow{3}{=}{Organize Desktop} &
        \multirow{3}{=}{Clear multiple desktop objects into their target containers.} &
        1 & The left hand places left-side objects, such as the banana and carrot, onto the plate. &
        \multirow{3}{=}{Multi-object sequencing, arm-specific selection, and subgoal tracking.} \\
        & & 2 & The right hand places the pen and tool into the basket. & \\
        & & 3 & The right hand places the plastic bag roll into the basket. & \\
        \midrule
        \multirow{3}{=}{Prepare Soy Milk} &
        \multirow{3}{=}{Prepare soy milk powder using a cup and spoon.} &
        1 & The right hand moves the cup to the workspace center. &
        \multirow{3}{=}{Fine-grained tool use, contact-rich manipulation, and bimanual coordination.} \\
        & & 2 & The left hand scoops soy milk powder with the spoon and pours it into the cup. & \\
        & & 3 & The left hand stirs the mixture and returns the spoon. & \\
        \bottomrule
    \end{tabular}
    \vspace{-6pt}
\end{table}

Together, these four tasks cover complementary real-world control challenges: rigid-object placement, deformable manipulation, long-horizon multi-object organization, and fine-grained tool use.
Figure~\ref{fig:task_exec_appendix} and Table~\ref{tab:real_world_scoring} therefore illustrate the task execution process and the basis for the 0--3 scoring criteria.

\section{Inference Speedup Details}
\label{app:inference_speedup}

\paragraph{Measurement protocol.}
We report $L_{\text{chunk}}$, the end-to-end latency of a single action-chunk inference call.
This includes image encoding, incremental planner-context access, OVCR context routing, and Action DiT denoising.
Video-DiT prefill runs asynchronously with action inference; we therefore use $L_{\text{chunk}}$ as the primary latency metric for closed-loop action frequency and separately track prefill throughput to ensure that planner context can be refreshed in the background.
All latency values are measured in milliseconds on RoboTwin 2.0 tasks unless otherwise specified.
The first warmup episode is discarded, and remaining per-episode means are averaged.
Both baseline and optimized pipelines use bf16 precision, so the reported speedups do not come from reducing numerical precision.

\begin{table}[!htbp]
    \centering
    \caption{Latency notation used in the inference speedup analysis.}
    \label{tab:latency_notation}
    \begin{tabular}{ll}
        \toprule
        Symbol & Definition \\
        \midrule
        $L_{\text{chunk}}$ & End-to-end latency of one action-chunk inference \\
        $L_{\text{prefill}}$ & Latency of one asynchronous Video-DiT prefill call \\
        \bottomrule
    \end{tabular}
\end{table}

\paragraph{CUDA acceleration.}
Table~\ref{tab:cuda_ablation} reports the cumulative optimization path from PyTorch eager inference to the optimized CUDA deployment.
The main deployment optimizations are grouped into three categories.
First, graph-level static capture compiles backbone modules including the Action DiT, memory/context modules, and VAE encoder into TensorRT engines and replays the fixed denoising loop with CUDA Graphs, reducing Python dispatch and redundant kernel launch overhead.
Second, selective \texttt{torch.compile} is applied to the Video-DiT prefill path, where full trace export is difficult due to Python-side control flow.
Third, hot-path redundancy elimination hoists chunk-level computations outside the denoising loop and skips repeated host-to-device copies for tensors that remain fixed across denoising steps.

\begin{table}[!htbp]
    \centering
    \caption{Cumulative CUDA acceleration ablation on RTX 5090D. CG denotes CUDA Graph.}
    \label{tab:cuda_ablation}
    \begin{tabular}{rlrrc}
        \toprule
        Stage & Optimization & $L_{\text{chunk}}$ (ms) & $L_{\text{prefill}}$ (ms) & Runtime \\
        \midrule
        0 & PyTorch eager baseline & $415.77 \pm 0.33$ & $61.15 \pm 0.26$ & A \\
        1 & + Action DiT TRT + CG & $83.87 \pm 0.51$ & -- & A \\
        2 & + Memory/context TRT + CG & $71.37 \pm 0.15$ & -- & A \\
        3 & + Video-DiT prefill \texttt{compile} & $71.45 \pm 0.14$ & $34.59 \pm 0.08$ & A \\
        4 & + Hot-path redundancy elimination & $50.37 \pm 0.27$ & -- & A \\
        5 & + Runtime upgrade (A $\to$ B) & $47.72 \pm 0.08$ & -- & B \\
        6 & + TRT video-KV skip-copy & $45.77 \pm 0.03$ & -- & B \\
        7 & + VAE encoder TRT & $41.37 \pm 0.03$ & -- & B \\
        \bottomrule
    \end{tabular}
    \vspace{3pt}
    \begin{flushleft}
    \footnotesize Runtime A: PyTorch 2.7.1 + cu128 / Triton 3.3.1 / TensorRT 10.16.1.11. Runtime B: PyTorch 2.12.0 + cu130 / Triton 3.7.0 / TensorRT 10.16.1.11.
    \end{flushleft}
\end{table}

\paragraph{Prefill compilation.}
The goal of prefill optimization is to make planner-context refresh available fast enough for asynchronous deployment.
Table~\ref{tab:prefill_compile} compares compile modes.
Although \texttt{reduce-overhead} gives the lowest prefill latency, it increases $L_{\text{chunk}}$; we therefore use the default mode.

\begin{table}[!htbp]
    \centering
    \caption{Video-DiT prefill compile mode comparison.}
    \label{tab:prefill_compile}
    \begin{tabular}{lrr}
        \toprule
        Configuration & $L_{\text{chunk}}$ (ms) & $L_{\text{prefill}}$ (ms) \\
        \midrule
        Eager (no compile) & 71.37 & $57.15 \pm 0.14$ \\
        compile (default) & 71.45 & $34.59 \pm 0.08$ \\
        compile (reduce-overhead) & 75.08 & $25.70 \pm 0.22$ \\
        \bottomrule
    \end{tabular}
\end{table}

\paragraph{Hot-path streamlining.}
The denoising loop contains computations that depend only on chunk-level inputs, such as condition embeddings, positional encodings, and references to planner-context K/V tensors.
Hoisting these computations outside the 10-step loop reduces $L_{\text{chunk}}$ from $71.45$ ms to $63.25$ ms.
Removing redundant recursive state traversals over modules already in evaluation mode further reduces latency to $50.37$ ms.
The TensorRT wrapper also reuses static input buffers for video-KV tensors that remain unchanged within a chunk, reducing latency by another $1.95$ ms in the 10-step setting.

\paragraph{VAE encoder validation.}
We validate the VAE encoder TensorRT engine against a PyTorch eager reference using both random inputs and real RoboTwin frames. As shown in Table~\ref{tab:vae_validation}, the TensorRT output remains numerically close to the reference and produces no NaN or Inf values.

\begin{table}[!htbp]
    \centering
    \caption{VAE encoder numerical validation against PyTorch eager reference.}
    \label{tab:vae_validation}
    \begin{tabular}{lrr}
        \toprule
        Metric & Random inputs ($N=1000$) & Real frames ($N=1000$) \\
        \midrule
        Cosine similarity (mean) & 0.999974 & 0.999941 \\
        Cosine similarity (min) & 0.999897 & 0.999589 \\
        Max absolute deviation (min--max) & 0.031--0.078 & 0.023--0.086 \\
        Relative max deviation (mean) & 1.24\% & 1.31\% \\
        NaN/Inf count & 0 & 0 \\
        \bottomrule
    \end{tabular}
\end{table}

\paragraph{ODE distillation.}
After CUDA acceleration establishes the optimized 10-step path, ODE distillation reduces the number of action denoising steps needed by the sampler.
For \modelname{}-Flash, we freeze the video DiT and distill only the action denoising path so that the student preserves the same planner-context and OVCR interface as the 10-step teacher.
The teacher produces a 16-step denoising trajectory, indexed from the noisy state $0$ to the final denoised state $16$, and we select trajectory anchors $\{0,1,2,4,8,12,16\}$.
During training, the student randomly samples a non-final anchor as the starting state and directly predicts the teacher's final denoised action state using a regression loss.
We sample more frequently from states near the noisy end of the trajectory, because accurate prediction from high-noise states is the key requirement for reducing the number of inference denoising steps.

\begin{table}[!htbp]
    \centering
    \caption{Latency of the CUDA-accelerated action sampler under different ODE-distilled denoising step counts.}
    \label{tab:ode_step_latency}
    \begin{tabular}{rrrr}
        \toprule
        Denoising steps & $L_{\text{chunk}}$ (ms) & Frequency (Hz) & Latency vs. 10-step \\
        \midrule
        1 & 14.67 & 68.3 & $-64.60\%$ \\
        2 & 17.56 & 56.9 & $-57.50\%$ \\
        4 & 23.45 & 42.6 & $-43.30\%$ \\
        10 & 41.37 & 24.2 & -- \\
        \bottomrule
    \end{tabular}
\end{table}

\clearpage

\section{Per-Task RoboTwin Success Rates}
\label{app:per_task_success}

Table~\ref{tab:per_task_success} reports per-task success rates on RoboTwin 2.0 for \modelname{} and the selected baselines.
All values are percentages, with clean and randomized evaluation reported separately.

\begingroup
\setlength{\LTcapwidth}{\linewidth}
\setlength{\tabcolsep}{2pt}
\renewcommand{\arraystretch}{0.85}
\begin{longtable}{@{}>{\scriptsize\raggedright\arraybackslash}p{0.18\linewidth}|*{2}{>{\scriptsize\centering\arraybackslash}p{0.055\linewidth}}|*{2}{>{\scriptsize\centering\arraybackslash}p{0.055\linewidth}}|*{2}{>{\scriptsize\centering\arraybackslash}p{0.055\linewidth}}|*{2}{>{\scriptsize\centering\arraybackslash}p{0.055\linewidth}}|*{2}{>{\scriptsize\centering\arraybackslash}p{0.055\linewidth}}|*{2}{>{\scriptsize\centering\arraybackslash}p{0.055\linewidth}}@{}}
    \caption{Per-task success rates on RoboTwin 2.0 under clean and randomized evaluation settings.}\label{tab:per_task_success}\\
    \toprule
    \multirow{2}{*}{Task} & \multicolumn{2}{c|}{\scriptsize \modelname{}} & \multicolumn{2}{c|}{\scriptsize \modelname{}-Flash} & \multicolumn{2}{c|}{\scriptsize Fast-WAM} & \multicolumn{2}{c|}{\scriptsize LingBot-VA} & \multicolumn{2}{c|}{\scriptsize $\pi_{0.5}$} & \multicolumn{2}{c}{\scriptsize Motus} \\
    \cmidrule(lr){2-3}\cmidrule(lr){4-5}\cmidrule(lr){6-7}\cmidrule(lr){8-9}\cmidrule(lr){10-11}\cmidrule(lr){12-13}
    & Clean & Rand. & Clean & Rand. & Clean & Rand. & Clean & Rand. & Clean & Rand. & Clean & Rand. \\
    \midrule
    \endfirsthead
    \toprule
    \multirow{2}{*}{Task} & \multicolumn{2}{c|}{\scriptsize \modelname{}} & \multicolumn{2}{c|}{\scriptsize \modelname{}-Flash} & \multicolumn{2}{c|}{\scriptsize Fast-WAM} & \multicolumn{2}{c|}{\scriptsize LingBot-VA} & \multicolumn{2}{c|}{\scriptsize $\pi_{0.5}$} & \multicolumn{2}{c}{\scriptsize Motus} \\
    \cmidrule(lr){2-3}\cmidrule(lr){4-5}\cmidrule(lr){6-7}\cmidrule(lr){8-9}\cmidrule(lr){10-11}\cmidrule(lr){12-13}
    & Clean & Rand. & Clean & Rand. & Clean & Rand. & Clean & Rand. & Clean & Rand. & Clean & Rand. \\
    \midrule
    \endhead
    \bottomrule
    \endfoot
    \bottomrule
    \endlastfoot
    Adjust Bottle & 100\% & 100\% & 100\% & 100\% & 100\% & 100\% & 90\% & 94\% & 100\% & 99\% & 89\% & 93\% \\
    Beat Block Hammer & 100\% & 94\% & 100\% & 93\% & 99\% & 97\% & 96\% & 98\% & 96\% & 93\% & 95\% & 88\% \\
    Blocks Ranking RGB & 100\% & 98\% & 100\% & 100\% & 100\% & 100\% & 99\% & 98\% & 92\% & 85\% & 99\% & 97\% \\
    Blocks Ranking Size & 98\% & 98\% & 97\% & 91\% & 94\% & 98\% & 94\% & 96\% & 49\% & 26\% & 75\% & 63\% \\
    Click Alarmclock & 100\% & 100\% & 100\% & 100\% & 100\% & 100\% & 99\% & 100\% & 98\% & 89\% & 100\% & 100\% \\
    Click Bell & 100\% & 100\% & 100\% & 100\% & 100\% & 100\% & 100\% & 100\% & 99\% & 66\% & 100\% & 100\% \\
    Dump Bin Bigbin & 99\% & 92\% & 95\% & 95\% & 97\% & 96\% & 89\% & 96\% & 92\% & 97\% & 95\% & 91\% \\
    Grab Roller & 100\% & 100\% & 100\% & 100\% & 100\% & 100\% & 100\% & 100\% & 100\% & 100\% & 100\% & 100\% \\
    Handover Block & 96\% & 85\% & 97\% & 87\% & 95\% & 81\% & 99\% & 78\% & 66\% & 57\% & 86\% & 73\% \\
    Handover Mic & 99\% & 99\% & 94\% & 92\% & 99\% & 100\% & 94\% & 96\% & 98\% & 97\% & 78\% & 63\% \\
    Hanging Mug & 85\% & 78\% & 67\% & 53\% & 58\% & 62\% & 40\% & 28\% & 18\% & 17\% & 38\% & 38\% \\
    Lift Pot & 100\% & 100\% & 100\% & 100\% & 100\% & 100\% & 100\% & 99\% & 96\% & 85\% & 96\% & 99\% \\
    Move Can Pot & 78\% & 83\% & 89\% & 94\% & 90\% & 88\% & 94\% & 97\% & 51\% & 55\% & 34\% & 74\% \\
    Move Pillbottle Pad & 100\% & 99\% & 96\% & 98\% & 100\% & 99\% & 99\% & 99\% & 84\% & 61\% & 93\% & 96\% \\
    Move Playingcard Away & 100\% & 100\% & 98\% & 100\% & 100\% & 100\% & 100\% & 99\% & 96\% & 84\% & 100\% & 96\% \\
    Move Stapler Pad & 88\% & 77\% & 75\% & 63\% & 77\% & 64\% & 91\% & 79\% & 56\% & 42\% & 83\% & 85\% \\
    Open Laptop & 98\% & 97\% & 98\% & 98\% & 98\% & 100\% & 92\% & 94\% & 90\% & 96\% & 95\% & 91\% \\
    Open Microwave & 40\% & 46\% & 9\% & 23\% & 62\% & 45\% & 82\% & 86\% & 34\% & 77\% & 95\% & 91\% \\
    Pick Diverse Bottles & 89\% & 85\% & 83\% & 87\% & 80\% & 85\% & 89\% & 82\% & 81\% & 71\% & 90\% & 91\% \\
    Pick Dual Bottles & 98\% & 99\% & 96\% & 98\% & 100\% & 96\% & 100\% & 99\% & 93\% & 63\% & 96\% & 90\% \\
    Place A2B Left & 100\% & 92\% & 96\% & 98\% & 95\% & 93\% & 97\% & 93\% & 87\% & 82\% & 88\% & 79\% \\
    Place A2B Right & 96\% & 92\% & 98\% & 96\% & 93\% & 99\% & 97\% & 95\% & 87\% & 84\% & 91\% & 87\% \\
    Place Bread Basket & 89\% & 94\% & 89\% & 98\% & 91\% & 93\% & 97\% & 95\% & 77\% & 64\% & 91\% & 94\% \\
    Place Bread Skillet & 91\% & 87\% & 87\% & 85\% & 90\% & 93\% & 95\% & 90\% & 85\% & 66\% & 86\% & 83\% \\
    Place Burger Fries & 99\% & 98\% & 94\% & 98\% & 96\% & 99\% & 97\% & 95\% & 94\% & 87\% & 98\% & 98\% \\
    Place Can Basket & 81\% & 94\% & 83\% & 79\% & 71\% & 69\% & 81\% & 84\% & 62\% & 62\% & 81\% & 76\% \\
    Place Cans Plasticbox & 100\% & 100\% & 98\% & 100\% & 99\% & 96\% & 100\% & 99\% & 94\% & 84\% & 98\% & 94\% \\
    Place Container Plate & 100\% & 96\% & 100\% & 94\% & 96\% & 100\% & 99\% & 97\% & 99\% & 95\% & 98\% & 99\% \\
    Place Dual Shoes & 91\% & 99\% & 92\% & 89\% & 94\% & 88\% & 94\% & 89\% & 75\% & 75\% & 93\% & 87\% \\
    Place Empty Cup & 100\% & 100\% & 100\% & 98\% & 100\% & 100\% & 100\% & 100\% & 100\% & 99\% & 99\% & 98\% \\
    Place Fan & 100\% & 87\% & 96\% & 94\% & 96\% & 96\% & 99\% & 93\% & 87\% & 85\% & 91\% & 87\% \\
    Place Mouse Pad & 93\% & 82\% & 79\% & 81\% & 83\% & 89\% & 93\% & 96\% & 60\% & 39\% & 66\% & 68\% \\
    Place Object Basket & 81\% & 80\% & 89\% & 85\% & 89\% & 88\% & 91\% & 88\% & 80\% & 76\% & 81\% & 87\% \\
    Place Object Scale & 96\% & 87\% & 89\% & 77\% & 90\% & 97\% & 96\% & 95\% & 86\% & 80\% & 88\% & 85\% \\
    Place Object Stand & 94\% & 93\% & 96\% & 90\% & 90\% & 94\% & 99\% & 96\% & 91\% & 85\% & 98\% & 97\% \\
    Place Phone Stand & 98\% & 96\% & 98\% & 94\% & 97\% & 99\% & 97\% & 97\% & 81\% & 81\% & 87\% & 86\% \\
    Place Shoe & 91\% & 96\% & 94\% & 98\% & 96\% & 99\% & 98\% & 98\% & 92\% & 93\% & 99\% & 97\% \\
    Press Stapler & 98\% & 99\% & 98\% & 100\% & 90\% & 97\% & 85\% & 82\% & 87\% & 83\% & 93\% & 98\% \\
    Put Bottles Dustbin & 83\% & 96\% & 83\% & 77\% & 95\% & 90\% & 87\% & 91\% & 84\% & 79\% & 81\% & 79\% \\
    Put Object Cabinet & 91\% & 89\% & 89\% & 85\% & 94\% & 89\% & 85\% & 87\% & 80\% & 79\% & 88\% & 71\% \\
    Rotate QRcode & 90\% & 90\% & 88\% & 90\% & 93\% & 89\% & 96\% & 91\% & 89\% & 87\% & 89\% & 73\% \\
    Scan Object & 94\% & 90\% & 85\% & 91\% & 89\% & 92\% & 96\% & 91\% & 72\% & 65\% & 67\% & 66\% \\
    Shake Bottle & 100\% & 100\% & 100\% & 100\% & 100\% & 100\% & 100\% & 97\% & 99\% & 97\% & 100\% & 97\% \\
    Shake Bottle Horizontally & 100\% & 100\% & 100\% & 100\% & 100\% & 100\% & 100\% & 99\% & 99\% & 99\% & 100\% & 98\% \\
    Stack Blocks Three & 99\% & 98\% & 100\% & 98\% & 95\% & 97\% & 99\% & 98\% & 91\% & 76\% & 91\% & 95\% \\
    Stack Blocks Two & 100\% & 99\% & 100\% & 100\% & 100\% & 100\% & 100\% & 98\% & 97\% & 100\% & 100\% & 98\% \\
    Stack Bowls Three & 79\% & 82\% & 81\% & 83\% & 80\% & 81\% & 86\% & 83\% & 77\% & 71\% & 79\% & 87\% \\
    Stack Bowls Two & 100\% & 100\% & 94\% & 98\% & 92\% & 98\% & 94\% & 98\% & 95\% & 96\% & 98\% & 98\% \\
    Stamp Seal & 98\% & 90\% & 81\% & 83\% & 90\% & 94\% & 96\% & 97\% & 79\% & 55\% & 93\% & 92\% \\
    Turn Switch & 70\% & 74\% & 53\% & 65\% & 61\% & 59\% & 44\% & 45\% & 62\% & 54\% & 84\% & 78\% \\
    \midrule
    Average & 93.40\% & 92.20\% & 90.48\% & 89.92\% & 91.88\% & 91.78\% & 92.90\% & 91.50\% & 82.74\% & 76.76\% & 88.66\% & 87.02\% \\
\end{longtable}
\endgroup

\clearpage

\end{document}